\def\year{2019}\relax
\documentclass[letterpaper]{article} 
\usepackage{aaai19}  
\nocopyright
\usepackage{times}  
\usepackage{helvet}  
\usepackage{courier}  
\usepackage{url}  
\usepackage{graphicx}  
\usepackage{latexsym}
\usepackage{amsmath}
\usepackage{amsmath,amsfonts,amssymb}
\usepackage{CJKutf8}
\usepackage{url}
\usepackage{algorithm}
\usepackage{algorithmic}
\usepackage{graphicx}
\usepackage{paralist, tabularx}
\usepackage{float}
\usepackage{multirow}
\usepackage{appendix}

\usepackage{tabularx}
\usepackage{color}

\newcommand{\tabincell}[2]{\begin{tabular}{@{}#1@{}}#2\end{tabular}}  
\title{A Hybrid Semantic Parsing Approach for Tabular Data Analysis}

\author{Yan Gao, Jian-Guang Lou, Dongmei Zhang \\
	Microsoft Research Asia, Beijing, China \\
	\{Yan.Gao, jlou, dongmeiz\}@microsoft.com \\
	2018.3.27
	}

\begin{document}
\maketitle
\begin{abstract}
This paper presents a novel approach to translating natural language questions to SQL queries for given tables, which meets three requirements as a real-world data analysis application: \textit{cross-domain}, \textit{multilingualism} and \textit{enabling quick-start}. 
Our proposed approach consists of:
(1) a novel data abstraction step before the parser to make parsing  table-agnosticism;
(2) a set of semantic rules for parsing abstracted data-analysis questions to intermediate logic forms as tree derivations to reduce the search space; 
(3) a neural-based model as a local scoring function on a span-based semantic parser for structured optimization and efficient inference.  
Experiments show that our approach outperforms state-of-the-art algorithms on a large open benchmark dataset WikiSQL. We also achieve promising results on a small dataset for more complex queries in both English and Chinese, which demonstrates our language expansion  and quick-start ability.
\end{abstract}

\section{Introduction}

\begin{figure}
	\centering
	\small
	\setlength{\abovecaptionskip}{0pt}
	\setlength{\belowcaptionskip}{-15pt}
	\includegraphics[width=5cm]{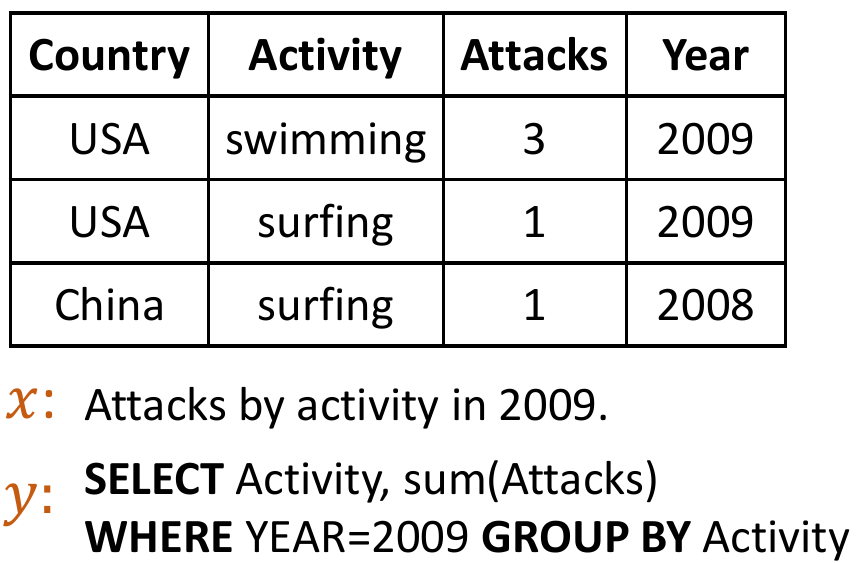}
	\caption{Give a table ``Shark Attacks'', $x$ is a question and $y$ is its corresponding SQL query. }
	\label{fig:introductionpic}
\end{figure}

Translating natural language questions to SQL queries is an important problem~\cite{gulwani2014nlyze,xu2017sqlnet,Iyer2017learning}.
As shown in Figure~\ref{fig:introductionpic}, our goal is to design an approach to translate a natural language question to its corresponding SQL query for a given table, which can be considered as a typical semantic parsing problem
~\cite{liang2011learning,lu2014semantic,pasupat2015compositional,cheng2017learning}.
As a real-world data-analysis application, there are three practical requirements.
First, end-uses may come from various fields.
When users upload a brand-new table from \textit{cross-domain}, it should work reasonably well without retraining the model.
Second, 
questions may be asked by different native speakers according to the table content (e.g., if the table content is Chinese, people tend to ask Chinese questions).
There is a clear evidence in usability research which shows that localization leads to improved user experience and greater client satisfaction, thereby promoting a need to support multilingual data-analysis interface. 
Third, \textit{enabling quick-start} is another desired properties we want. 
Collecting data is costly and time-consuming in reality, especially for language expansion.
The less effort on composing hand-crafted features or collecting labeled training data that is used, the better the process is.
\begin{figure*}
	\centering
	\setlength{\abovecaptionskip}{0pt}
	\setlength{\belowcaptionskip}{-15pt}
	\includegraphics{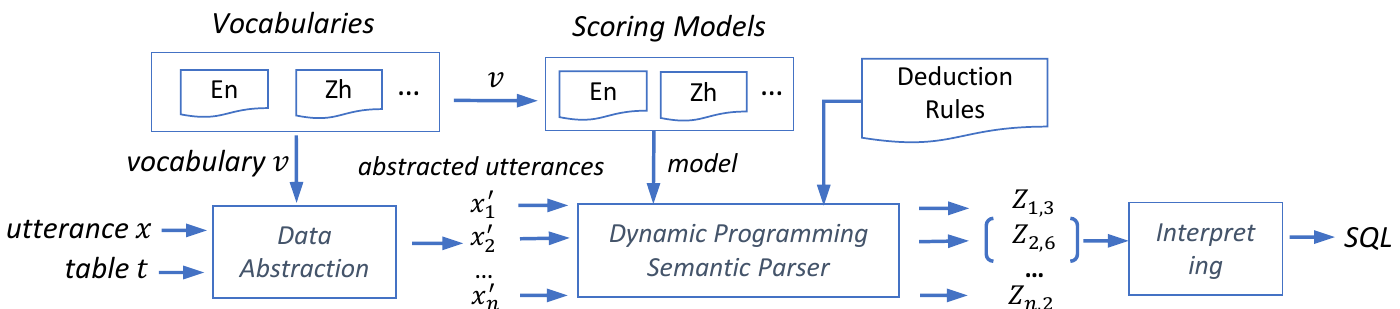}
	\caption{Online framework of our approach. ``En'' means English and ``Zh'' means Chinese.}
	\label{fig:workflow}
\end{figure*}

However it is hard to meet the three requirements at the same time.
Traditional text-to-SQL approaches~\cite{Pazos2013NLIDB,li2014nalir,Li2016Understanding} of Database community rely heavily on the syntactic structure of a sentence, which bring intrinsic complexity for language expansion;
Existing multilingual semantic parsers~\cite{Susanto2017NeuralAF,Duong2017MultilingualSP} are restricted at specific domain;
Recent neural semantic parsing models~\cite{zhong2017seq2sql,yu2018typesql} need a lot of training data to generalize to new tables.
However questions often contain rare entities (e.g. domain-specific technical terms) and numbers/dates specific to the underlying data tables.
Most of such key words have out-of-vocabulary (OOV) problem for those word embedding models in cross-domain and multilingual scenario.
Existing methods~\cite{Pinter2017MimickingWE,Bahdanau2017LearningTC} use a more complex encoder to build representation for rare words on-the-fly from subword units.
It will add additional runtime and reduce online efficiency, especially for big tables, that are not preferred by industries in reality.

In this paper, we argue that the content of entities should not reflect semantic parsing, but their data type (e.g. string, number or date) and symbol type (e.g. column name or entry value) play a key role, because entities with similar meta-data information can be arbitrarily substituted in users' questions (e.g. "2009" can be replaced with "2008" given table in Figure~\ref{fig:introductionpic}).
With this motivation, we first use a novel data abstraction step before parsing to identify table related entities mentioned in questions and abstract them with their meta-data information, and then feed the abstracted questions into parser. In this way, our parsing model is table-agnostic and can be applied to cross-domain and multilingual scenario. In addition, it largely reduces the noise of data and enhances the knowledge shared between those abstracted questions, and thus boosts the performance of parer and enables a \textit{quick-start} with small data set.


In semantic parsing, deduction rules and model are the most important parts, that largely influence the expressiveness and accuracy of the parser 
~\cite{liang2016learning}.
Our principle is designing a set of purely semantic rules, and depending on neural model to learn the mapping between surface syntax-structures and deduction rules. 
Devising deduction rules may need efforts but worth-well since it helps to reduce the search space and increase the interpretability. 
Moreover it is a one-short deal since all rules are universal cross languages and data tables.


In this paper, we propose a hybrid semantic parsing approach for text-to-SQL task to meet three requirements: \textit{cross-domain}, \textit{multilingualism} and \textit{enabling quick-start}. 
The main contributions of our approach are as follows:
\begin{compactitem}
	\item We use a novel data abstraction step before parsing to identify table related entities mentioned in questions, which makes our parser table-agnosticism. (Section~\ref{section:data abstraction})
	\item We propose a set of deduction rules to generate compositional logic forms as tree derivations for data-analysis questions. 
	The proposed deduction rules are universal cross languages and tables, and avoid searching in a huge space. (Section \ref{sec_grammar})
    \item We use a neural-based local scoring function on a span-based semantic parser for structured global optimization and efficient inference.
   	Our neural-based model builds soft alignments between surface structures of questions and deduction rules as local scoring functions, achieving \textit{quick-start} without using any hand-crafted or language-specific features.  (Section \ref{section:model})
    \item We evaluate our approach and achieve state-of-the-art accuracy on a large open benchmark dataset (i.e., WikiSQL).
    In addition, we also obtain preliminary but promising results on a small dataset consisting of more complex queries in two languages. (Section \ref{sec_evaluation})
\end{compactitem}

\section{Approach}
\label{sec_approach}
In this section, we present the overall workflow of our approach.
Figure~\ref{fig:workflow} illustrates the online steps of our approach:
1) given an utterance $x$ and a table $t$,  
and a vocabulary $v$, we perform a \textit{data abstraction} step to map $x$ to multiple abstracted utterances $\{x_1',\dots x_n'\}$, which makes parser table-agnostic.
2) for each abstracted utterance $x_i'$, we run a bottom-up dynamic programming parser with a set of deduction rules and a neural local scoring model to obtain the highest scored intermediate logic form $Z_{i,j}$;
3) at last, we select the logic form with top score and interpret it to a SQL query.
In our framework, each language has its own vocabulary and  scoring model. While other modules are universal cross languages.
Details about our approach are presented in this section.


%

\begin{figure}[t]
	\centering
	\setlength{\abovecaptionskip}{0pt}
	\setlength{\belowcaptionskip}{-20pt}
	\includegraphics[width=0.8\linewidth]{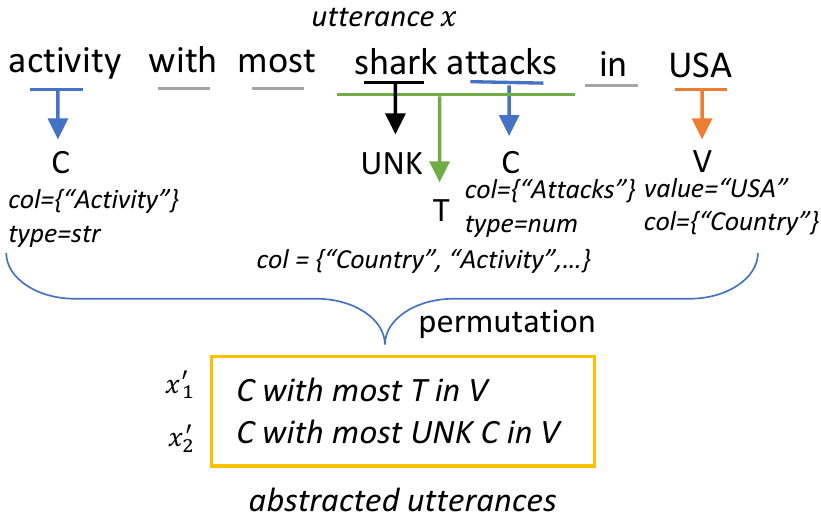}
	\caption{Data abstraction maps utterance $x$
		to a set of abstracted utterances $\{x_1',\dots x_n'\}$.}
	\label{fig:dataabstractionpic}
\end{figure}

\subsection{Data Abstraction}
\label{section:data abstraction}

At first, a language detection module is used to detect the language of utterance $x$ and select corresponding vocabulary $v$ and tokenization method\footnote{In our product, each language has its own tokenization module.}.
After that, as shown in Figure~\ref{fig:dataabstractionpic}, data abstraction maps $x$ to multiple sequences of tokens in $v$ (denoted as abstracted utterances $\{x_1',\dots x_n'\}$), and initialize their properties. We will first introduce the vocabulary, and then the mapping methods.
\subsubsection{Vocabulary}
Our vocabulary consists of 3 parts:
\begin{compactitem}
	\item some data-analysis common words collected from 4 aspects: a). important stopping words (e.g., by, of); b). aggregation words (e.g., sum, average); c). operation words (e.g., group, sort); d). comparison words (e.g., top, most). 
	
	\item 5 meta-data symbols (e.g., $T$, $C$, $V$, $N$ and $D$) defined in Table \ref{tab:table schema}, which represent basic entities in data-analysis area. Each symbol has a set of properties describing their information.  For example, $col$ records the set of column names that a symbol contains; $type$$\in$$\{str,date,num\}$ records the data type of a column (C); $value$ stores the string value of the  table entry ($V$), the number ($N$), or the date/time ($D$). 
	The functions of symbol properties are two-fold: 
	1) instantiate meta-data symbols during \textit{interpreting} step for converting intermediate logic forms to final SQL queries.
	2) support precondition checking of deduction rules during \textit{parsing} step. 
	\item a special token ``UNK'' indicates unknown noisy words.
\end{compactitem}

Each language has its own vocabulary.
There are overall 400 tokens in our English vocabulary and 130 tokens in our Chinese vocabulary.
The vocabulary is small because (1) all data-related words are abstracted to meta-data symbols; (2) to reduce the noise of semantic parsing, only a small set of data-analysis common words are picked. 

\subsubsection{Mapping}
Recognizing common words in $v$ is straightforward by stemming and exact string matching.
While replacing table-related words with meta-data symbols is a non-trivial task.
For example, the mapping is not unique since words might be mapped to different symbols (e.g., as shown in Figure~\ref{fig:dataabstractionpic}, ``shark attacks'' can be mapped to \textit{T}, and ``attacks'' can be mapped to \textit{C}, since \textit{Shark Attacks} is a table name and \textit{Attacks} is a column name of that table). 

\begin{table}[t]
	\small
	\centering
	\setlength{\abovecaptionskip}{5pt}
	\setlength{\belowcaptionskip}{-5pt}
	\begin{tabular}{ccc}
		\textbf{Symbol} & \textbf{Properties} & \textbf{Semantic}  \\ \hline
		$T$ & \textit{col} & table\\
		$C$ & \textit{col, type} & column \\
		$V$ & \textit{value, col} & table entry\\
		$N$ & \textit{value, col} & number \\ 
		$D$ & \textit{value, col} & date/time \\
		\hline
	\end{tabular}
	\caption{Meta-data symbol, property and their semantics}
	\small 
	\label{tab:table schema}
	\vspace{-10pt}
\end{table}
In order to better support these complex situations, we allow some words to have multiple overlapping annotations. 
In the example of Figure~\ref{fig:dataabstractionpic}, ``shark attacks'' can be mapped to either ``UNK+C'' or a single ``T''. 
Thus, two abstracted utterances ($x'_1$ and $x'_2$) are obtained in this case.
Our model will finally address the ambiguity by ranking all possible logic forms and selecting the one with the highest score. 

In our approach, in order to meet the low latency requirement of the real scenarios (especially for a big table), we currently use a rule based NER system for data abstraction\footnote{In real scenarios, we provide real-time auto-complete suggestions to help users avoid typos and inconsistencies between terms and table entities to facilitate our NER system.}.
For example, we use Recognizer\footnote{https://github.com/Microsoft/Recognizers-Text} to recognize numbers (N) and date/time (D) in $x$, which supports multiple languages.
For other symbols, given an utterance $x$ and a table $t$, 
we first normalize all words
in $x$ and $t$
with NLTK\footnote{http://www.nltk.org/api/nltk.stem.html} that supports lemmatization for 17 languages. 
Next, we tokenize $x$ into \textit{n}-grams of length from 1 to 5 and use them to match meta-data in the $t$. 
For table name (T), column name (C) and entry value (V), firstly, we iterate all \textit{n}-grams by exact string matching. 
If there is no matching, we look for the \textit{n}-gram with highest edit-distance score that is larger than a predefined threshold (set as 0.8). 
If there is still no  matching, we launch a synonym dictionary from project X (Anonymized for Blined Review) to find possible synonym \textit{n}-grams for that meta-data. 
Similar techniques are also applied in other systems~\cite{gao2015datatone,dhamdhere2017analyza}. 
Afterwards, for those words without any annotation, we mark them as ``UNK''. 
Finally, we use a standard permutation algorithm to generated a set of non-overlapping token sequences.

Data abstraction brings several advantages:
(1) it makes parser table-agnostic; (2) 
With a small vocabulary, it largely reduces the noise of data and enhances the knowledge shared between abstracted questions, and thus boosts the performance of parer and enables a \textit{quick-start} with small data set.

\begin{table}[t]
	\small
	\centering
	\setlength{\abovecaptionskip}{5pt}
	\setlength{\belowcaptionskip}{-5pt}
	\begin{tabular}{ccc}
		\textbf{Symbol} & \textbf{Properties} & \textbf{Semantic}  \\ \hline
		$A$ & \textit{col} & aggregation\\
		$G$ & \textit{col} & grouping by\\
		$F$ & \textit{col} & filtering \\
		$S$ & \textit{col} & superlative \\ \hline
	\end{tabular}
	\caption{Operator symbol, property and their semantics}
	\small 
	\label{tab:operator_symbol}
	\vspace{-10pt}
\end{table}



\subsection{Semantic Parsing}
\label{section:parsing}
Given abstracted utterances, instead of directly synthesizing SQL queries, we first generate an intermediate semantic representation with a set of deduction rules, and then further convert it to SQL query.
The semantic representation is a kind of logic form that is restricted as tree structure, which is preferred for statistical semantic parsing, because it can be constructed from a sentence compositionally~\cite{liang2013lambda}. 

\subsubsection{Deduction Rules}
\label{sec_grammar}
\begin{table*}
	
	\fontsize{10}{\baselineskip}
	\centering
	\setlength{\abovecaptionskip}{5pt}
	\setlength{\belowcaptionskip}{-15pt}
	\small
	\begin{tabular}{lcc}
		$Symbols\rightarrow \hat{Symbol}:[predicate]$ & $Precondition$ & $Schema$  \\ \hline
		\multicolumn{3}{c}{\underline{Composition Rule}} \\ 
		$C|A|G|S+T\rightarrow\hat{T} $: $[project]$ & $(C|A|G|S).col\subseteq T.col$& $\hat{T}.col=(C|A|G|S).col$ \\	
		$F+T\rightarrow\hat{T} $: $[filter]$ & $F.col \subseteq T.col$& $\hat{T}.col=T.col$ \\	
		$A|G+C\rightarrow\hat{G}$:$[group]$ & $C.type \subseteq \{str, date\}$ & {$\hat{G}.col =(A|G).col \cup C.col$ } \\
		$C+V|D\rightarrow\hat{F}$:$[equal]$  &  $C.col=(V|D).col$ & $\hat{F}.col=C.col$ \\ 
		$C+N\rightarrow\hat{F}$:$[more|less|>=| <=]$ & $C.col=N.col$ & $\hat{F}.col=C.col$ \\ 
		$F+F\rightarrow\hat{F}$:$[and| or]$ & N/A & $\hat{F}.col=F.col \cup F.col$ \\
		$A|C_1+C_2\rightarrow\hat{S}$:$[argmax| argmin]$ & $C.type \in \{str, date\}$, $ C_2.type=num$ &  $\hat{S}.col=A|C_1.col\cup C_2.col$ \\ 
		$C+C\rightarrow\hat{C}$:$[combine]$  &  N/A & $\hat{C}.col=C.col \cup C.col$\\	
		$A+A\rightarrow\hat{A}$:$[combine]$  &  N/A & $\hat{A}.col=A.col \cup A.col$\\		
		$C|A|G|S$+$F$$\rightarrow$$\hat{C}|\hat{A}|\hat{G}|\hat{S}$:$[modify]$ & $(C|A|G|S).col\cap F.col=\emptyset$  & $(\hat{C}|\hat{A}|\hat{G}|\hat{S}).col=(C|A|G|S).col$\\
		\multicolumn{3}{c}{\underline{Raising Rule}} \\
		$C\rightarrow\hat{A}$:$[min|max|sum|avg]$ & $C.type=num$ & $\hat{A}.col=C.col$ \\
		$C\rightarrow\hat{A}$:$[count]$ & $C.type \in \{str, date\}$ & $\hat{A}.col=C.col$ \\
		$V|D\rightarrow\hat{F}$:$[equal]$ & N/A  &$\hat{F}.col=(V|D).col$ \\  
		\hline	
		
	\end{tabular}
	\caption{Full set of our deduction rules.}
	\label{tab:Grammar_table}
\end{table*}

In order to support multilingual scenario,
our deduction rules are purely semantic rules that only use symbols.
Language-related tokens are taken as span contexts.
In this way, all rules are universal cross languages, 
and we depend on model to learn different syntax structures of different languages. 
In addition, all rules can be applied to new tables since there is no lexical rules.
Table~\ref{tab:Grammar_table} shows the full set of deduction rules we use. 
Each rule is a triple: 
\begin{compactitem}
	\item $Symbols$$\rightarrow$$\hat{Symbol}$:$[predicates]$ means a new symbol can be deduced from left symbols and generate corresponding predicates. 
	Except 5 meta-data symbols defined in Table~\ref{tab:table schema}, we have 4 operator symbols defined in Table ~\ref{tab:operator_symbol}, which indicate high-order data-analysis operations (e.g., filter, aggregate, group and superlative). Operator symbols and their properties are deduced from meta-data symbols with corresponding predicates.
	Those predicates are mainly derived from relational algebra.
	
	\item $Preconditions$ are used to prune invalid or redundant rules to reduce the search space during parsing, which are mainly derived from our prior knowledge in data analysis domain. 
	For example, a string-type column can only do ``count'' aggregation; Superlative operation should be conducted on a numerical column and return a string-value column. 
	We use deduction rules with 9 kinds of preconditions to represent such knowledge. 
	Notably, these preconditions are universal cross languages and data tables. 
	\item $Schema$ defines how to set the properties of a newly deduced symbol when we apply a rule during parsing.
\end{compactitem}
Take the first deduction rule in Table~\ref{tab:Grammar_table} as an example:
a \textit{project} operation can be conducted to a table $T$ and a column $C$, then produce a new table $\hat{T}$, only if $T$ contains the columns specified by $C$.  
Finally, the columns of $\hat{T}$ are assigned as $\hat{T}.col=C.col$.

We have two kinds of deduction rules: \textit{composition rules} and \textit{raising rules}.
\textit{Composition rules} compose two symbols into a new one, which are the most basic components of our rule system as it helps to generate tree structure logic forms.
Note that the symbol order at the left side does not matter (e.g., ``$C + T$'' and ``$T + C$'' are the same), which makes the parser more robust to questions with ill-formed syntactic structures (e.g., key-word based queries). 

There are some types of divergence between syntactic and semantic scopes caused by certain synthetic variations. For example, in
``\textit{sales of BMW which is more than 3000}'', table entry ``\textit{BMW}'' is in the middle of column name ``\textit{sale}'' and number ``\textit{3000}'', which will stop their compose.
To maintain the compositionality, we design a \textit{modification mechanism} to handle this problem. 



The \textit{modification mechanism} is inspired by the X-bar theory in constituency~\cite{jackendoff1977x,chomsky2002syntactic}, 
in which phrases can be headed by a word with some modifiers (e.g., $NP$:=$NP$+$PP$).
Similarity, in our rule system, filter symbol $F$ can be combined with other symbols as a modifier
(e.g., $C|A|G|S$+$F$$\rightarrow$$\hat{C}|\hat{A}|\hat{G}|\hat{S}$:$[modify]$), 
only if they contain disjointed columns  (e.g., $(C|A|G|S).col$$\cap$$F.col$=$\emptyset$).
In modification rule, the newly deduced symbol type keeps the same with the left \textit{head} symbol type, and inherits all its properties.
With such a modifying operation, we can keep attaching more $F$ onto a \textit{head} symbol
and align semantic composition to a linguistic head-modifier phrasing structure to parallel constituency parsing.





Composition rules assume that only two adjacent symbols can produce a new symbol,
which is not flexible enough to handle implicit intentions.
It is very common that there are missing tokens in utterances that reflect important semantics.
For example, in \textit{``shark attacks by country''}, the aggregation intent $sum$ is implicit from the context; in \textit{``sales of bmw''}, column name ``brand'' is also  omitted.
To allow our semantic representation to handle implicit intentions,
we propose one-to-one \textit{raising rules}.
For example, $C$$\rightarrow$$\hat{A}$:$[min|max|sum|avg]$ allows a symbol $A$ to be deduced from $C$ with different predicates. Also, a single $V$ can directly produce a filter operation with predicate \textit{equal} on its column (e.g., $V$$\rightarrow$$\hat{F}$:$[equal]$ and $\hat{F}.col$=$V.col$). 
\subsubsection{Parsing Algorithm}
In order to enable a fast  span-based dynamic programming parser with our deduction rules, we extend the standard CYK~\cite{Chappelier1998AGC} parser.
Given an abstracted utterance, we take the meta-data symbols as leaf nodes, and take common tokens and ``UNK'' as the span context of tree nodes. The semantic parse tree ($Z$) is obtained by applying the deduction rules in Table~\ref{tab:Grammar_table} of the following two kinds: 
\begin{align*} 
\scriptsize
(X, s)[\iota]&\rightarrow(\hat{X}, s)[\hat{\iota}]  \\
(X_1, s_1)[\iota_1]+(X_2, s_2)[\iota_2]&\rightarrow \nonumber(\hat{X}, s_1\Join s_2)[\hat{\iota}]
\end{align*}
The first rule represents a \textit{raising rule} that
takes symbol $X$ with token span $s$ and produces symbol $\hat{X}$ with predicate $\hat{\iota}$ on the same span. 
The second rule is a \textit{compositional rule} that takes two adjacent symbols (i.e., $X_1$ and $X_2$) and create a new symbol $\hat{X}$ with predicate $\hat{\iota}$ on the conjunction of their spans $s_1 \Join s_2$. 
The initial span context of a leaf node is itself.
Each non-leaf tree node in $Z$ is a tuple ($r$, $s$), where $r$ is the applied deduction rule in Table~\ref{tab:Grammar_table} and $s$ indicates the corresponding span that triggers $r$.



\subsubsection{Scoring Model}
\label{section:model}
\begin{figure}[t]
	\centering
	\setlength{\abovecaptionskip}{0pt}
	\setlength{\belowcaptionskip}{-15pt}
	\includegraphics[width=0.7\columnwidth]{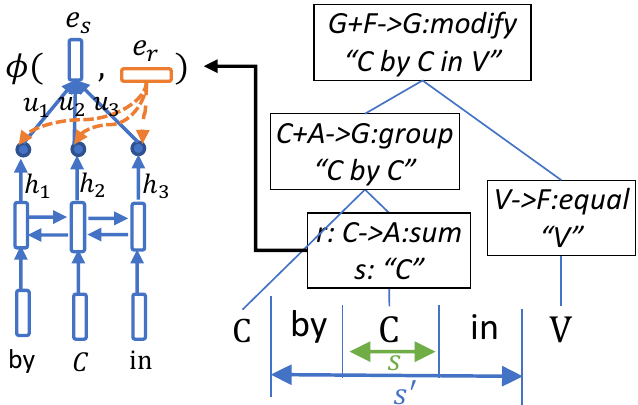}
	\caption{Each non-leaf node ($r$, $s$) in the semantic tree is scored independently  by a neural function $\phi$.
	}
	\label{fig:model_pic}
\end{figure}

The overall score  of tree $Z$ is integrated by independent scores of each non-leaf tree node $z$,
which enables efficient dynamic programming during inference.
Inspired by recently works in chart-based neural parser \cite{durrett2015neural,stern2017minimal}, 
 we utilize a neural model to learn the mapping between surface structures and deduction rules as our scoring function. 
For each node, not only tokens within the span, but also tokens outside it also provide a rich source of information~\cite{Hall-Durrett-Klein:2014:SpanParser} without using any hand-crafted or language-specific features. 
In order to capture more useful information and avoid too much noise,  we expand the span bidirectionally until a symbol occurs as the surface structure for each applied deduction rule.
For example, in Figure~\ref{fig:model_pic}, for node ($C$$\rightarrow$$\hat{A}$:$[sum]$, ``\textit{C}''), the surface structure $s'$ is obtained by expanding the span ``\textit{C}'' until symbols ``\textit{C}'' and ``\textit{V}'' are reached.

As shown in Figure~\ref{fig:model_pic}, for each non-leaf node ($r$, $s$), we first embed  $r$ into a dense vector ($e_r = Wf_r$) with a matrix $W$ and a sparse feature vector $f_r\in \{0,1\}^d$ of $r$. Currently, we only use the rule's identity as $f_r$, while other rule features (e.g., right symbol identity) can also be easily incorporated in our framework.
Next,  each token in our vocabulary is represented as an embed vector. Given a surface structure $s'$=$(x{'}^{(1)},x'^{(2)},\dots x'^{({n})})$,
we first obtain a hidden state sequence $(h_1,\dots h_n)$ by a bidirectional LSTM network.
An attention mechanism is then adopted to further transform $(h_1,\dots h_n)$ to a final representation $e_s$ as follows:
\begin{align*} 
\small
u_i &=\theta^\intercal\tanh(W_1h_i+W_2e_{r}) \\
a_i &=\frac{\exp(u_i){}}{\sum_{j=1}^n \exp(u_j)} \\
e_s &=\sum_{i=1}^{n} a_{i}h_i
\end{align*}
The vector $\theta$ and token embeddings and matrices $W$, $W_1$ and $W_2$ are all trainable model parameters.

Finally, the probability of a tree $Z$ is calculated through a log-linear distribution:
\[\label{eq:1} p(Z|x)\propto \exp(\sum_{z_i=(r_i,s_i)\in Z} \hspace{-10pt} \phi(e_{r_i},e_{s_i}))\]
where $z_i$ is a non-leaf tree node and $\phi$ can be any scoring function(e.g. dot product or cosine similarity).
We use dot product in this paper for simplicity in practice.
With our span-based parser and neural scoring model, we can get highest scored $Z$ with dynamic programming inference.


\section{Training Process}

Different from online dynamic programming inference, during training, for each labeled question-SQL pair given table $t_i$ (denoted as $(x_i, t_i, y_i)$), our semantic parser generates all valid semantic parse tree ($Z$) candidates.
For each valid $Z$, we further convert it to an SQL query and compare it against the ground truth $y_i$. 
The conversion is  straightforward. 
We traverse tree $Z$ in pre-ordering and use a set of heuristic rules to map each node to different SQL clauses. We ignore the details because it is not a key technical problem, and we submitted our code along with this paper as a supporting material.
If two SQL queries are equivalent, the logic form is considered a consistent logic form. Otherwise, it is inconsistent. Here, we use canonical SQL matching, which ignores the ordering difference of elements in different SQL clauses.
 
\subsection{Training objective}
Given training data $\{(x_i, t_i, y_i)\}_{i=1}^N$, we try to minimize the following pair-wise loss function:
\[L=\sum_{i=1}^{N} max(0, \alpha-\sum_{j=1}^{M}\sum_{k=1}^{K}(p(z_j^+|x_i)- p(z_k^-|x_i))\]
Here, $\{z_j^+, j=1\dots M\}$ are consistent logical forms of $x_i$ and 
$\{z_k^-, k=1\dots K\}$ are inconsistent logical forms  of $x_i$
and $\alpha$ is a margin parameter (set as $0.5$). 

\subsection{Pruning Search Space}
Essentially, the span-based parser will search all possible paths according to deduction rules.
A huge search space is a big problem for answering complex queries \cite{zhang2017macro}. 	
In order to reduce the search space, we incorporate rich prior knowledge in data analysis domain during parsing: 
(1) we carefully design our deduction rule system and use loop detection algorithm to avoid rule loops.
(2) we add preconditions for deduction rules to void invalid logic forms and empty results; For example, for each utterance on WikiSQL, the average number of valid semantic parse trees is only 33.8 .







\section{Evaluation}
\label{sec_evaluation}



To validate \textit{cross-domain}, \textit{multilingualism} and \textit{quick-start} abilities, we evaluate our approach on two Question-to-SQL datasets: WikiSQL~\cite{zhong2017seq2sql} and our newly proposed multilingual dataset (abbreviated as CommQuery for convenience).

\subsection{Methods for Comparison}
In this experiment, we compare our algorithm with state-of-the-art algorithms on WikiSQL: \textit{Seq2SQL} \cite{zhong2017seq2sql}, \textit{SQLNet} \cite{xu2017sqlnet},  \textit{STAMP}~\cite{sun2018semantic}, \textit{Corse2Fine}~\cite{Dong2018CoarsetoFineDF}, \textit{MQAN}~\cite{McCann2018TheNL}  and \textit{TypeSQL}~\cite{yu2018typesql}.
In order to compare with other deduction rule based approaches, 
we use float parser~\cite{pasupat2015compositional} with our deduction rules on WikiSQL (denoted as ``FloatParser''). 
\textit{Seq2SQL}, \textit{SQLNet}, \textit{Corse2Fine} and \textit{MQAN} don’t use table content.
\textit{STAMP}, \textit{TypeSQL}, FloatParser and our approach use table content. 
\textit{MQAN} uses extrac data, and \textit{TypeSQL} use extra knowledge.  
Here, our approach is mainly compared to table content aware methods.

In addition, for a better understanding of 
our design choices,
we also test some variants of our approach.
(1) \textit{data abstraction:} we use a common seq-to-sql model, where our abstracted utterance is encoded using a BiLSTM network with attention mechanism, and then decoded to SQL (denoted as ``Seq2SQL+abs'').
\textit{(2) surface structure:} we tried different surface structures,
including only inside tokens of $s$  (``+inside''), expanding left side of $s$ until a symbol occurs (``+left''), or expanding right side of $s$ until a symbol occurs (``+right'').
\textit{(3) scoring model:}  
``Our Approach+sparse'' indicates that the learning process is based on sparse features instead of neural features. Thus the probability of $Z$ is 
$ p(Z|x)\propto \exp\left(\sum_{z_i=(r_i, s_i)\in Z}  f_{s_i}^\top Wf_{r_{i}}\right)$,
where $f_{s_i}$ is a \textit{3}-gram vector extracted over each expanded span ($s_i'$) and $W$ is trainable parameter.
\subsection{Experimental Setup}

In our approach, each token in vocabulary is embedded into a 100
dimensional vector. Both the rule embedding and
the hidden states have a size of 50. We use
Adam~\cite{kingma2014adam} with a learning rate
0.001. The initial hidden state of BiLSTM is set to
be zero.
 



\subsection{Experiment Results on WikiSQL} 
\label{experimental results on WikiSQL}

%
%
%
%
WikiSQL is the largest public text-to-SQL dataset, including 56,324/8,421/15,878 examples in the training/dev/test set, all of which do not share tables. It can help to evaluate whether our approach can generalize to new tables or not.
We compare the canonical SQL matching accuracy, which ignores the ordering difference of elements in the \textit{Where} clause ($Acc_{qm}$) and execution results accuracy ($Acc_{ex}$) of different approaches. Results are shown in Table~\ref{tab:wikiSQL overall result}. 
Compared with \textit{FloatParser}, our span-based parser has a better accuracy, since it more effectively learns the mapping between surface structures and deduction rules based on span constraints. 
Compared with other neural-based methods on WikiSQL, our approach performs best,
since we use \textit{data abstraction} to make parser table-agnostic and combine neural network with a set of deduction rules to reduce the search space. For each utterance on WikiSQL, the average number of valid semantic parse trees is only 33.8 .  
\begin{table}[t]
	\setlength{\abovecaptionskip}{0pt}
	\setlength{\belowcaptionskip}{0pt}
	\small
	\begin{tabular}{>{\centering\arraybackslash}p{3.6cm}>{\centering\arraybackslash}p{0.7cm}>{\centering\arraybackslash}p{0.7cm}>{\centering\arraybackslash}p{0.7cm}>{\centering\arraybackslash}p{0.7cm}}
		& \multicolumn{2}{c}{Dev} & \multicolumn{2}{c}{Test} \\ \hline
		Model     & $Acc_{qm}$  &$Acc_{ex}$  & $Acc_{qm}$  & $Acc_{ex}$ \\ \hline 
		\textsc{Seq2SQL}~\cite{zhong2017seq2sql} &  - &60.8\%& - &59.4\%\\
		\textsc{SQLNet}~\cite{xu2017sqlnet}    & 63.2\% &  69.8\%& 61.3\%&68.0\% \\
		\textsc{FloatParser}    & 70.7\% & 73.1\%  &69.8\% &72.4\% \\
		\textsc{STAMP}~\cite{sun2018semantic}    & - &75.1\%  &- & 74.6\% \\
		\textsc{Corse2Fine}~\cite{Dong2018CoarsetoFineDF}  &-& 79.0\%& -  &78.5\% \\ 
		\textsc{MQAN}~\cite{McCann2018TheNL}  & -   &82\% & - & 81.4\%\\ 
		\textsc{TypeSQL}~\cite{yu2018typesql}  & 79.2\% & 85.5\%  &75.4\% &82.6\% \\ \hline
		Our Approach + inside    & 64.8\%$^\ast$	& 68.4\%$^\ast$  & 60.5\%$^\ast$ 	& 64.2\%$^\ast$ \\
		Our Approach + right    & 67.5\%$^\ast$ 	& 71.0\%$^\ast$ & 63.1\%$^\ast$ 	& 67.8\%$^\ast$ \\
		Our Approach + left    & 77.4\%$^\ast$  	& 81.7\%$^\ast$ & 75.2\%$^\ast$ 	& 80.5\%$^\ast$ \\
		Seq2SQL + abs	&78.1\%$^\ast$	& 81.8\%$^\ast$  &77.7\%$^\ast$ &81.0\%$^\ast$ \\
		Our Approach + sparse   & 79.3\%$^\ast$  & 82.8\%$^\ast$ &79\%$^\ast$ &82.5\%$^\ast$  \\
		\textbf{Our Approach} & \textbf{80.2\%}$^\ast$ &\textbf{84.2\%}$^\ast$   &\textbf{79.9\%}$^\ast$  & \textbf{84\%}$^\ast$ \\ \hline
	\end{tabular}
	\caption{Overall results on WikiSQL. $Acc_{qm}$ and $Acc_{exe}$ denote the accuracies of canonical representation and execution result respectively. ``-'' denotes that they don't report the canonical representation of SQL matching accuracy. ``$^\ast$'' denotes that the p-value $<$ 0.05. }
	\label{tab:wikiSQL overall result}
\end{table} 

\begin{table}[t]
	\centering
	\small
	\renewcommand{\multirowsetup}{\centering}
	\begin{tabular}{ccc}
		$Acc_{qm}$   & Error & Reason   \\ \hline
		\tabincell{c}{92.2\%   } &  4.7\%    & 	data abstraction errors \\
		in \textit{Select Column} & 3.1\% & incorrectly labeled SQLs \\ \hline
		\multirow{2}{2cm}{87.9\% \\ in \textit{Select Aggregator}} & 3.4\%  & mismatch sum and count    \\
		
		& 1.2\%      & data abstraction errors                \\
		& 7.4\%      & incorrectly labeled SQLs     \\ \hline
		\tabincell{c}{ 95.2\% } & 1.4\%      &\tabincell{c}{ data abstraction errors}     \\
		in \textit{Where Clause}	& 3.4\%      & incorrectly labeled SQLs      \\ \hline                                            
	\end{tabular}
	\setlength{\abovecaptionskip}{0pt}
	\setlength{\belowcaptionskip}{-10pt}
	\caption{Fine-grained error analysis over WikiSQL.}
	\label{error analysis}
\end{table}

\begin{figure}[t]
	\centering
	\setlength{\abovecaptionskip}{0pt}
	\setlength{\belowcaptionskip}{0pt}	
	\includegraphics[width=0.6\columnwidth]{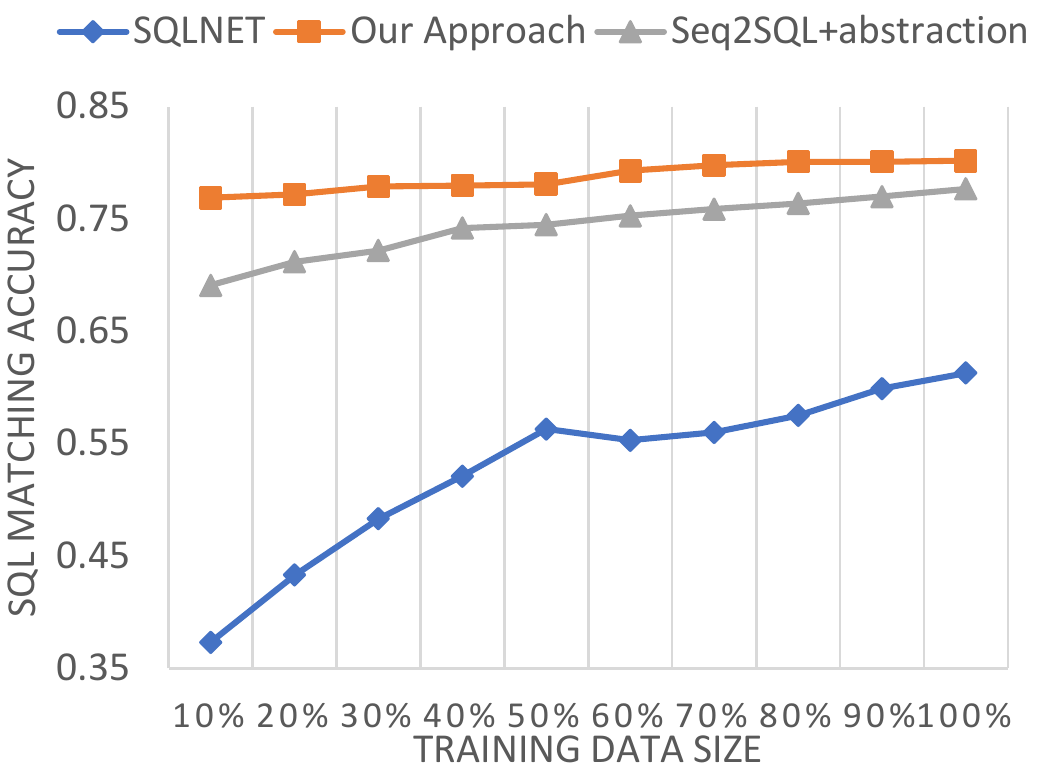}
	\caption{$Acc_{test}$ vs. training data size on WikiSQL.}
	\label{fig:experiment_pic}
\end{figure}


From Table~\ref{tab:wikiSQL overall result}, we can conclude which parts of our approach that are actually better than some counterparts. Take $Acc_{qm}$ on test set as an example:
(1) among four different surface structures, ``Our Approach+inside'' performs worst since it contains the least information. ``+left'' outperforms ``+right'' largely, indicating that most useful informations are at left side. Furthermore, expanding the span bidirectionally is the best choice;
(2) ``Seq2SQL+abs'' outperforms Seq2SQL 29.4\%, which indicates that \textit{data abstraction} can significantly boost the accuracy of a parser;
(3) our approach with neural features outperforms sparse features 2\% on the test set, which indicates that the neural network can capture richer non-linear features than sparse features.


To better understand the source of erroneous results of our approach, we report more fine-grained $Acc_{qm}$ and error ratios on the test set in ``\textit{Select Column}'', ``\textit{Select Aggregator}'' and ``\textit{Where Clause}''. As we can see in Table \ref{error analysis}, 
a significant portion of failed cases are caused by \textit{data abstraction} (e.g., it accounts for 4.7\%, 1.2\%, and 1.4\% in ``\textit{Select Column}'', ``\textit{Select Aggregator}'' and ``\textit{Where Clause}'' respectively). Those questions need common sense reasoning (e.g., ``people lived'' refers to column ``population'' and ``attended the game'' refers to column ``crowd''), which go beyond the capability of our current NER system. 


At last we further compare $Acc_{qm}$ on test when we change the size of training data from 10\% (5635 pairs) to 100\% (56355 pairs), but still keep the original dev and test set. 
As shown in Figure \ref{fig:experiment_pic}, 
the less training data we use, the larger the superiority \textit{Sql2SQL+abs} have with \textit{SQLNet}, which indicates that \textit{data abstraction} can largely reduce the volume of data that the neural model need. 
Also, compared with \textit{Sql2SQL+abs},
\textit{Our Approach} has more siginificant improvement on smaller training data (e.g., improves 9.1\% and 2.3\% on 10\% and 100\% training data respectively), because deduction rules can further reduce the search space of parsing.  It demonstrates the \textit{quick-start} ability of our approach with small training data set.
We don't report the result of TYPESQL and MQAN since they need extra resources.

\subsection{Experiment Results on CommQuery}
\begin{table}[t]
	\centering
	\small
	\setlength{\abovecaptionskip}{0pt}
	\setlength{\belowcaptionskip}{0pt}
	
	\begin{tabular}{c|c}
		\hline Type  & Example  \\ \hline
		Statistic
		  & please compute sum of sales  for me\\
		Group by & show me sales for each brand and category    \\
		Superlative 
		& top selling brands for each year  \\
		Filter  & show me sales of BMW by each category        \\
		Comparison  & compare sales of BMW and Toyota              \\
		Pinpoint   & select brands whose sale is more than 30000  \\ \hline
	\end{tabular}
	\caption{\small Basic question types and examples in CommQuery}
	\label{tab:Question_Types_and_Examples_in_CommQuery}
\end{table}
\begin{table}[t]
	
	\small
	\setlength{\abovecaptionskip}{0pt}
	\setlength{\belowcaptionskip}{0pt}
	\centering
	\begin{tabularx}{7.5cm}{XXXX>{\centering\arraybackslash}X>{\centering\arraybackslash}X>{\centering\arraybackslash}X}
		\centering 	
		 &Total&S&SW&SG&SWG&SWGH \\ \hline
		Wiki &80,654& 0.8\%& 99.2\%& -& - & - \\ 
		En&1,005& 2.9\%& 62.8\%& 12.1\%& 14\%&8.2\% \\
		Zh&918&5.8\%&26.7\%&21.9\%&28.6\%&18.4\% \\     \hline          
	\end{tabularx}
	\caption{\small Contribution of SQL forms in WikiSQL and CommQuery, where S,W,G,H represent different SQL clauses (e.g., \textit{SELECT}, \textit{WHERE}, \textit{GROUP BY} and \textit{HAVING} clauses). A character set means multiple clauses coexist. For example, ``SG'' means SQLs that only contain \textit{SELECT} and \textit{GROUP BY} clauses.}
	\label{tab:statistics of complex query}
\end{table}
\begin{table}[t]
	\small
	\centering
	\setlength{\abovecaptionskip}{0pt}
	\setlength{\belowcaptionskip}{-15pt}
	\begin{tabular}{ccc}
		\textbf{Model}& \textbf{English}    & \textbf{Chinese}    \\ \hline
		 Seq2SQL + abs  & 81.5\%$^\ast$  & 74.2\%$^\ast$  \\
		Our Approach + inside    & 86.8\%$^\ast$  & 87.9\%$^\ast$  \\
		Our Approach + right      & 88.5\%$^\ast$   & 88.6\%$^\ast$   \\

		Our Approach + sparse & 92.1\%$^\ast$   & 91.8\%$^\ast$ \\
		Our Approach + left    & 95.4\%$^\ast$  & 94.2\%$^\ast$  \\
		\textbf{Our Approach} & \textbf{96.4\%}$^\ast$ & \textbf{95.1\%}$^\ast$ \\ \hline
	\end{tabular}
	\caption{Overall SQL canonical representation accuracies on CommQuery. ``$^\ast$'' denotes that the p-value $<$ 0.05.}
	\label{tab:overall result in complex queries}
\end{table}

However, WikiSQL has some intrinsic limitations that make it different from real-world scenario: (1) SQLs in WikiSQL dataset are quite simple, only with one column or one aggregator in the $Select$ clause, and no $Groupby$ clause.  (2) most questions and table content are in English. (3) questions always contain exact entry values in tables.  
To validate the ability of handling more complex queries in real-world scenarios, we construct 3 tables in both English and Chinese from 3 public tables (e.g.,  \textit{SharkAttacks} (https://bit.ly/2oIr8Xi),
\textit{Jobs} (https://bit.ly/2LYMAAj) and
\textit{CarSales} (https://bit.ly/2MMuhUq)), and 
then collect 1,005 English and 918 Chinese question-SQL pairs on them
as a new benchmark (CommQuery). 
These question-SQL pairs are provided by our vendors. We provide six basic question types and examples for them (Table~\ref{tab:Question_Types_and_Examples_in_CommQuery}), and
they can ask questions by freely combining these basic types.
To compare these real-world queries with WikiSQL queries, we categorize them into various SQL forms by the  kinds of SQL clauses it contains. As is shown in Table~\ref{tab:statistics of complex query}, queries in CommQuery are much richer than WikiSQL in terms of the distribution of SQL forms.
All approaches are trained on 15 epochs and the SQL matching accuracies are tested with 5-fold cross validation.

The canonical SQL matching accuracies are shown in Table~\ref{tab:overall result in complex queries}.
We make four observations: 
(1) Our approach outperforms ``Seq2SQL + abs'' by 14.9\% in English and 20.5\% in Chinese, which is more significant than the 2.3\% improvement on WikiSQL. 
Our approach has more significant improvement for complex queries than those of simple queries in WikiSQL.
(2) Expanding the span bidirectionally as surface structure is still the best choice for more complex queries.
(3) Even in a small but complex queries dataset, our neural features can still outperform sparse features, with very promising overall result (96.4\% in English and 95.1\% in Chinese)
which will enable a quick-start in real world scenarios.
(4) We extend our approach from English to Chinese by only constructing a Chinese vocabulary with 130 tokens and building a scoring model with 918 question/SQL pairs, which demonstrates our easy language expansion ability.




\section{Related Work and Discussion}
\label{sec_relatedwork}

Semantic parsing maps natural language to executable programs.
For building question answering systems, semantic parsing has emerged as an important and powerful paradigm ~\cite{berant2013semantic,pasupat2015compositional,liang2016learning,zhang2017macro}.  
Parsers define deduction rules based on grammar formalism such as Combinatory Categorial Grammar (CCG) ~\cite{zettlemoyer2007online,kwiatkowski2010inducing,kwiatkowski2011lexical,krishnamurthy2013jointly,kushman2013using},
Synchronous CFG~\cite{wong2007learning}, 
and CFG~\cite{kate2006using,chen2011learning,berant2013semantic,desai2016program}; 
Others use the syntactic structure of the utterance to guide the composition ~\cite{poon2009unsupervised,reddy2016transforming,reddy2017universal}. 
In this paper, we propose a set of CFG rules for parsing data-analysis questions to intermediate logic forms as tree derivations for text-to-SQL task.
Inspired by recent works about neural chart parsing \cite{durrett2015neural,stern2017minimal}, 
we use a neural-based model as local scoring function on a span-based semantic parser to avoid hand-crafted features.



Recent neural based semantic parsing approaches~\cite{Iyer2017learning,zhong2017seq2sql} achieved promising results on text-to-SQL task. 
Some of previous works assume that user queries contain exact string entries in the table. However, it is unrealistic that users always formulate their questions with exact string entries in the table. For example, being unfamiliar with table content often leads to vocabulary gap in natural language (e.g., user queries ``US'' but the entry value is ``USA'' ). This makes generating the right SQLs without searching the database content in such cases impossible. 
To tackle these problems, similar with TypeSQL~\cite{yu2018typesql}, in this work, we target at content-sensitivity text-to-SQL scenario.

Identifying table entities mentioned in questions is a critical subproblem of semantic parsing. Existing methods take it as a part of semantic parsing through grammar rule~\cite{liang2011learning,pasupat2015compositional}, or as an entity linking sub-network in an encoder-decoder framework \cite{krishnamurthy2017neural}. 
Other works in semantic parsing for text-to-SQL task decouple semantic parsing and entity linking as two separate stages. 
Corse2Fine~\cite{Dong2018CoarsetoFineDF} uses the first decoder to generate a semantic sketch, and then uses the second decoder to fill in missing details.
SQLizer~\cite{yaghmazadeh2017type} first uses SEMPRE~\cite{berant2013semantic} to generate SQL sketch, and then uses program synthesis techniques and iteratively repair if necessary. 
Conversely, we first use a data abstraction step to generate abstracted utterances, and then feed them into parser to generate final SQL queries.  







\section{Conclusion}
\label{sec_conclusion}

In this paper, we propose a hybrid semantic parsing approach for data-analysis scenario to meet three practical requirements: \textit{cross-domain}, \textit{multilingualism} and \textit{enabling quick-start}. 
We achieve state-of-the-art results on a large open benchmark dataset WikiSQL. We also achieve promising results on a small dataset for more complex queries in both English and Chinese, which demonstrates our language expansion and quick-start ability.


\bibliographystyle{aaai}
\bibliography{aaai}

\end{document}